\documentclass[10pt, a4paper]{article}

\usepackage[final]{lrec-coling2024} % this is the new style

\usepackage{arydshln}
\usepackage{caption}
\usepackage{subcaption}

\usepackage{multirow}

\usepackage{enumitem}

\usepackage{amsmath}
\usepackage{booktabs}

\usepackage{pifont}% http://ctan.org/pkg/pifont
\newcommand{\heart}{\ensuremath\heartsuit}

\title{Contextualizing Generated Citation Texts}

\name{Biswadip Mandal, Xiangci Li, Jessica Ouyang} 

\address{The University of Texas at Dallas \\
         Richardson, TX, USA \\
         \{biswadip.mandal, xiangci.li, jessica.ouyang\}@utdallas.edu\\
         }

\abstract{Abstractive citation text generation is usually framed as an infilling task, where a sequence-to-sequence model is trained to generate a citation given a reference paper and the context window around the target; the generated citation should be a brief discussion of the reference paper as it relates to the citing context. However, examining a recent LED-based citation generation system, we find that many of the generated citations are generic summaries of the reference paper's main contribution, ignoring the citation context's focus on a different topic. To address this problem, we propose a simple modification to the citation text generation task: the generation target is not only the citation itself, but the entire context window, including the target citation. This approach can be easily applied to any abstractive citation generation system, and our experimental results show that training in this way is preferred by human readers and allows the generation model to make use of contextual clues about what topic to discuss and what stance to take.}

\begin{document}
\maketitleabstract

\section{Introduction}

Citation text generation is the task of generating a short summary of a reference paper, focused on how it relates to a citing paper. Modern approaches \cite{xing2020automatic, ge-etal-2021-baco, luu-etal-2021-explaining,chen-etal-2021-capturing,li-etal-2022-corwa} frame this problem as an infilling task: a context window of up to three sentences before and after the target citation is extracted, and the target is masked; given this context window and the reference paper abstract, a sequence-to-sequence (seq2seq) model is trained to generate the masked target citation.

If we examine the context window of a target citation, we find citations of similar related works, and the author may criticize them in order to draw attention to a gap that the citing paper intends to fill; to a human reader, the context window provides clues indicating the topic the citation will discuss and the stance the author will take towards the reference paper. Thus, a citation should not be a generic summary of the reference paper, but rather a query-focused summary, based on the context. After all, different papers can cite the same reference paper in completely different ways, depending on whether they share a task, an approach, or a dataset; even within the same citing paper, a reference paper can be cited differently in the introduction versus in the methodology or comparison of results.

It has been observed that neural language models often generate text that is generic and vague \cite{holtzman2019curious}. Examining the recent LED-based citation generation system of \citet{li-etal-2022-corwa}, we see a similar problem. Figure \ref{fig:para_example_specific} shows an example of their LED model's output. Although the generated citation is a valid summary of the reference paper, it is too generic, mentioning only ``syntactic information." The context sentence containing the masked target already describes a constituent-structure approach, making it clear to a human reader that the target should discuss a corresponding dependency-structure based approach. However, the LED model produces a generic summary of the reference paper, which while factually correct, does not fit the context well and is actually redundant with information given earlier in the context. 

\begin{figure}
\small
    \begin{tabular}{l}
    \toprule
    \vspace{.5\baselineskip}
    
    \begin{minipage}[h]{0.9\columnwidth}%
    \textbf{Context:}
    Broadly speaking, prior work on SRL makes use of syntactic information in two different ways. Carreras and Marquez (2005); Pradhan et al. (2013) incorporate constituent-structure span-based information, while \textbf{[MASK]}\ldots \\

	\textbf{Ground truth:} Haji et al. (2009) incorporate dependency-structure information.\\

	\textbf{LED-baseline:} Haji et al. (2009) integrate syntactic information into a neural SRL system.\\ 
	
	\textbf{Contextualized:} Haji et al. (2009) use dependency-level information.\\
	\end{minipage}\tabularnewline
	\bottomrule
    \end{tabular}
    \caption{An infilling-style baseline approach produces a generic summary of the reference paper that does not fit well in the citation context.}
    \label{fig:para_example_specific}
\end{figure}

To address this issue, we propose \textit{contextualized citation generation}, a context-focused modification to the citation text generation task. In this version of the task, the model must generate not only the target citation, but the surrounding context as well. In this way, the context window serves as a sort of prompt for the decoder. Figure \ref{fig:para_example_specific} shows how the output of our contextualized approach correctly identifies dependency parsing as the topic of the citation, and our experimental results show that human readers prefer citations generated using our approach over the baseline infilling approach. Our proposed training method can be easily applied to any seq2seq citation generation model to produce coherent citations that fit their contexts.

%We use the CORWA dataset of related work sections from natural language processing papers \cite{li-etal-2022-corwa} to investigate the quality of the generation and its fit in the related work context. When generating a citation span for a given related work context paragraph, the output should be coherent with the context. 

\begin{figure*}[!ht]
    \centering
    \includegraphics[width=.9\textwidth, height=.35\textwidth]{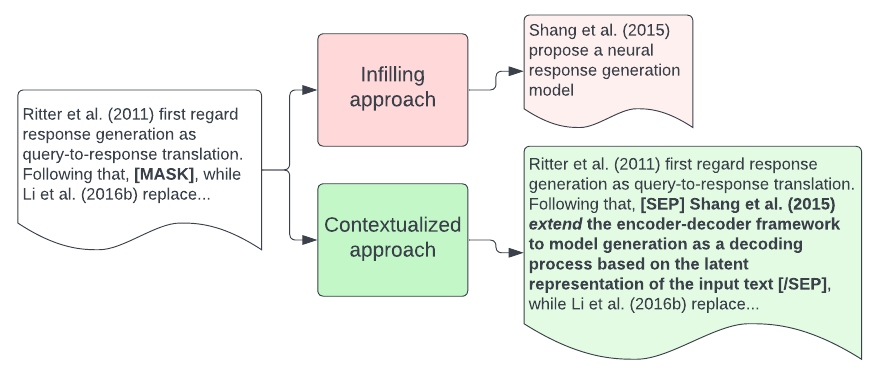}
    \caption{Comparison of the standard infilling-based citation text generation approach, where the generation target is just the target citation itself, and our proposed contextualized approach, where the generation target is the entire context window with the target citation filled in.}
    \label{fig:approach}
\end{figure*}

\section{Related Work}

The citation text generation task was proposed by \citet{hoang2010towards}. They and other extractive approaches \cite{hu-wan-2014-automatic, chen2019automatic} selected the most salient sentences from the reference papers to serve as their citations. 

More recently, neural abstractive approaches \cite{xing2020automatic, ge-etal-2021-baco, luu-etal-2021-explaining,chen-etal-2021-capturing,li-etal-2022-corwa} have trained seq2seq models for the citation infilling task, where the input is the citation context (with target citation masked) and the reference paper abstract. Our work aims to condition the generated citation even more strongly on the context by training the model to generate the entire context window, not just the target citation. While the above works differ in features and datasets used, as well as the size and shape of the context window, it is important to note that our approach changes only the generation target, and so is compatible with all of them. In our experiments, we use the most recent approach, \citet{li-etal-2022-corwa}, as the baseline model.

\section{Methodology}
\label{lb:para_gen}

We propose a simple modification of the citation text generation task in order to produce citations that are more coherent with their surrounding contexts. The task is to generate the entire context window, with the masked target citation filled in. For ease of evaluation, we add a meta-token {\tt [SEP]} to distinguish the context from the target citation. Figure \ref{fig:approach} illustrates the difference between our approach and the standard infilling approach.

\subsection{Experimental Setup}

\paragraph{Data.} We use the NLP-domain CORWA citation text generation dataset \cite{li-etal-2022-corwa}. Table \ref{tab:corwa_stats} shows the size of the dataset and its partitions. We merge the human-annotated \textit{train} and automatically-labeled \textit{distant} sets to create a single large, unified training set. 

\begin{table}[t]
\small
\centering
\begin{tabular}{lcc}
\toprule
\textbf{Partition} & \textbf{Papers} & \textbf{Citations} \\
\midrule
\textit{train} & 565 & 2,243 \\
\textit{distant} & 11,564 & 32,512 \\
\textit{test} & 362 & 1,322 \\
\bottomrule
\end{tabular}
\caption{CORWA dataset partitions and statistics.}
\label{tab:corwa_stats}
\end{table}

\paragraph{Model.} We use the Longformer Encoder-Decoder \citep[LED;][]{Beltagy2020Longformer} citation generation model of Li et al. The input to the model is the concatenation of the citing paper's introduction section; the paragraph containing the (masked) target citation; and the citation mark (eg. ``Smith et al. (2023)"), title, and abstract of the reference paper.

\section{Results and Analysis}

%\begin{table}[t]
%    \small
%    \centering
%    \begin{tabular}{lccc}
%    \toprule
%        Model & R-1 & R-2& R-L\\
%    \midrule
%        LED-Baseline & 0.271 & 0.075 & 0.208 \\ 
%        Contextualized & 0.254 & 0.065 & 0.194 \\ 
%    \bottomrule
%    \end{tabular}
%    \caption{Performance of the paragraph generation model in ROUGE F1.\\} 
%    \label{tab:para_generation_results}
%\end{table}

% rouge doesn't capture the contextual generation fit 
%As shown in Table \ref{tab:para_generation_results}, the ROUGE score for paragraph generation is lower than baseline span generation. However, the slight difference in ROUGE might not indicate a significant difference in generation quality, as studied in \cite{deutsch2022re}. It is also worth noting that, while Table \ref{tab:para_generation_results} shows only ROUGE F1 scores, we do observe a significant increase in ROUGE recall score and corresponding decrease in in precision, which may be due to the paragraph model generating longer spans (Section \ref{sec:longer_para_generation}).

%Additionally, ROUGE is based on word overlap statistics and does not capture other aspects of the summary, such as factuality \cite{wallace2021generating} or coherence. It can be misleading if used as the only measure to assess the informativeness of summaries \cite{schluter2017limits}. Hence, we perform a human evaluation of the generated spans, along with an in-depth analysis.

\subsection{Human Evaluation}
\label{subc:para_human_eval}

We recruit six graduate students from our university's Computer Science Department to serve as judges. We split the them into two groups of three, assigning each group 30 samples. The judges are shown the input to the generation model and the ground truth, baseline, and contextualized citations; the order of citations is randomly shuffled to anonymize the models. The judges are asked to indicate which citation they prefer based on \textit{Fluency}, \textit{Relevance} to the reference paper, \textit{Coherence} in the citation context, and \textit{Overall} quality. 

Table \ref{tab:para_generation_human_eval} shows our human evaluation results. We see that the proposed contextualized generation model is slightly preferred with respect to Relevance, Coherence, and Overall, while the baseline model is more fluent. Comparing the contextualized model with the ground truth, we find that many samples have indistinguishable (ie. comparable) performance. Interestingly, we observe that the Relevance of ground truth is judged lower than that of the contextualized model. This may be because the ground truth citations are written by human authors who have access to the entire reference paper, while the generation models and human judges only see the reference paper abstracts.

We achieve moderate inter-annotator agreement for the first group of three judges, with pairwise Kendall's $\tau$ of 0.31, 0.17, and 0.35. We had lower agreement for the second group, with Kendall's $\tau$ of 0.15, 0.08, and 0.04. Examining the judges' scores, we find that one judge from the second group consistently disagreed with the other two; we also tallied the results excluding this third judge, but we did not find any significant difference from the results shown in Table \ref{tab:para_generation_human_eval}.

\begin{table}[!t]
    \small
    \centering
    \begin{tabular}{lcccc}
    \toprule
        Model & Flu. & Rel. & Coh. & Overall\\
    \midrule
        \multicolumn{5}{c}{Contextualized vs Baseline} \\
    \midrule
        \heart Contextualized & 12 & \textbf{49} & \textbf{41} & \textbf{53}  \\
        \heart Baseline & \textbf{33} & 44 & 36 & 50 \\
        Indistinguishable & 135 & 87 & 103 & 77 \\
        
        % \multicolumn{5}{|c|}{Paragraph Generation vs Span Generation} \\ \hline
        % Paragraph Generation wins & 12(6.6\%) & \textbf{49(27.2\%)} & \textbf{41(22.7\%)} & \textbf{53(29.4\%)}  \\ \hline
        % Span Generation wins & \textbf{33(18.3\%)} & 44(24.4\%) & 36(20.0\%) & 50(27.7\%) \\ \hline
        % Indistinguishable & 135(75.0\%) & 87(48.3\%) & 103(57.2\%) & 77(42.7\%) \\ \hline
        
    \midrule
        \multicolumn{5}{c}{Contexualized vs Ground Truth} \\
    \midrule
        \heart Contextualized & \textbf{19} & \textbf{38} & \textbf{28} & \textbf{27}  \\ 
        \heart Ground Truth & 18 & 13 & 21 & 18 \\
        Indistinguishable & 143 & 129 & 131 & 135 \\
    \bottomrule
    \end{tabular}
    \caption{Human evaluation comparing contextualized and infilling-based citation generation.} \label{tab:para_generation_human_eval}
\end{table}

% JJO: Add some discussion of ground truth comparison -- done

% report the rouge correlation with human annotation as well. 

\subsection{Analyzing Contextualized Citations}
\label{subc:analyze_para_generation}

%Even though there is no fixed evaluation metric to evaluate the contextual fit, 
Examining the generated citations, we find several common scenarios where the baseline model ignores clues in the context window that the contextualized model is able use. %As part of the evaluation of the paragraph generation summaries, we illustrate such different scenarios with examples where the paragraph generation models paid attention to context. 

\begin{figure}
    \small
    \begin{tabular}{l}
    \toprule
    \vspace{.5\baselineskip}
    
    \begin{minipage}[h]{0.9\columnwidth}%
    \textbf{Context:} \ldots To the best of our knowledge, \textbf{[MASK]}. Most similar to our work in spirit, Ding et al. (1997) used Layer-wise Relevance Propagation (LRP; Bach et al al. 2015), \textit{an interpretation method resembling saliency}, to interpret \ldots \\

	\textbf{Ground truth:} Li et al. (2016) presented \textit{the only work that directly employs saliency methods} to interpret NLP models.\\

	\textbf{Baseline:} Li et al. (2016) proposed three strategies for visualizing compositionality in NMT.\\ 
	
	\textbf{Contextualized:} Li et al (2016) is \textit{the only work that applied saliency} to NMT. They used a simple recurrent neural network (RNN) to learn a saliency function for each word in the sentence.\\
	\end{minipage}\tabularnewline
	\bottomrule
    \end{tabular}
    \caption{Example where the context hints the citation should discuss saliency (in \textit{italics}).}
    \label{tab:para_example_generic}
\end{figure}

\paragraph{The context discusses the historical development of a particular topic.} A coherent citation should focus on the part of the reference paper that is related to that topic; even if that topic is not especially important in the reference paper, a good citation generation model should still be able to identify it as the topic of focus. Figure \ref{tab:para_example_generic} shows an example where the context paragraph contains two clues. First, the introductory phrase ``to the best of our knowledge" hints that the citation should claim that few, if any prior works have addressed the topic, identifying a gap that the citing paper intends to fill. Second, the following sentences states that the most similar work uses a ``method resembling saliency." The contextualized model is able to put these clues together, highlighting how ``Li et al (2016) is the only work that applied saliency."    

\begin{figure}
    \small
    \begin{tabular}{l}
    \toprule
    \vspace{.5\baselineskip}
    
    \begin{minipage}[h]{0.9\columnwidth}%
    \textbf{Context:} \ldots Previous work on \textbf{[MASK]}. \textit{In contrast with their work,} we train models to \ldots \\

	\textbf{Ground truth:} curriculum learning for MT (Kocmi and Bojar, 2017; Wang et al., 2018) proposed methods which feed easier samples to the model first and later show more complex sentences. \textit{However,} their focus is on improving convergence time while providing limited success on improving translation quality.\\

	\textbf{Baseline:} dynamic sentence sampling (Kocmi and Bojar, 2017; Wang et al., 2018) focuses on how to dynamically sample sentences in order to accelerate the training of NMT.\\ 
	
	\textbf{Contextualized:} fine-tuning neural machine translation models (Kocmi and Bojar, 2017; Wang et al., 2018) has tried to dynamically sample sentences in order to accelerate training. \textit{However,} these methods require training models to learn a prior distribution of similar sentences.\\
	\end{minipage}\tabularnewline
	\bottomrule
    \end{tabular}
    \caption{Example where the context hints the citation should criticize the reference paper (in \textit{italics}).}
    \label{tab:para_example_compare_one}
\end{figure}
    
\paragraph{The context implies that the reference paper should be compared or contrasted.} \citet{teufel2006automatic} identifies several author intents for citing a given paper, including contrasting methods or results, claiming the cited work is worse than or comparable to the current work, or comparing or contrasting with other cited works. A common pattern is that the context sentence following the masked target discusses how the current work addresses limitations in the prior work, making it evident that the target citation criticizes the reference paper. % While sometimes it is easy to infer if the masked span should carry a function of comparison/contrast, other time it may not be obvious. 
Figure \ref{tab:para_example_compare_one} %and \ref{tab:para_example_compare_three} 
shows an example where the context sentence immediately following the target citation begins with ``in contrast with their work."
    
    %\item\label{item:specific} \textbf{The citation span should go into specifics of the cited paper.} There are instances where the generated spans are valid summaries of the cited paper, but they are not specific enough. For example, the reason for citing the paper might be a specific method, but the baseline generated span only gives a generic overview of the cited paper. Table \ref{tab:para_example_specific} describes one such example.

\begin{figure}
    \small
    \begin{tabular}{l}
    \toprule
    \vspace{.5\baselineskip}
    
    \begin{minipage}[h]{0.9\columnwidth}%
    \textbf{Context:} \ldots Rubinstein et al. (2015) demonstrated that \textit{state-of-the-art distributional semantic models fail to predict attributive properties of concept words} (e.g. the properties is-red and is-round for the word apple) as accurately as taxonomic properties (e.g. is-a-fruit). \textit{Similarly}, \textbf{[MASK]} \ldots \\ %Collell and Moens (2016) compared linguistic and visual representations of object concepts on their ability to represent different types of property knowledge\ldots \\

	\textbf{Ground truth:} Sommerauer and Fokkens (2018) investigated the \textit{types of semantic knowledge encoded within pretrained word embeddings, concluding that some properties cannot be learned} by supervised classifiers.\\

	\textbf{Baseline:} Somasomauer and Fokkens (2018) compared word embeddings to a feature-identification method based on cosine similarity. \\ %They found that the feature-based similarity outperformed the log-linear model.\\ 
	
	\textbf{Contextualized:} Sommerauer and Fokkens (2018) showed that the \textit{properties of concepts are not captured by word embeddings}.\\
	\end{minipage}\tabularnewline
	\bottomrule
    \end{tabular}
    \caption{Example where the context hints the citation should discuss how the reference paper is similar to a previous citation (in \textit{italics}).}
    \label{tab:para_example_similar_two}
\end{figure}

\paragraph{The context implies that the reference paper should be discussed in a similar way as another work.} Figure \ref{tab:para_example_similar_two} shows an example where the sentence containing the target citation begins with ``similarly," hinting that the citation should discuss how the reference paper is similar to the citation in the previous sentence. The baseline model again produces a generic summary, while the contextualized model is able to identify the topic of focus as the failure of word embeddings to capture information about concept words.
    
\paragraph{The target citation should not be redundant with the context.} We find several instances where the baseline model's generated citation repeats information already present in the context. While the citation is still a valid summary of the reference paper, the baseline model fails to recognize that the information is redundant and it should focus on something else, such as a different method or concept. %These scenarios can be similar to ones in \ref{item:specific}, however they may or may not be very generic. 
The example in Figure \ref{fig:para_example_specific} shows a generic, redundant citation generated by the baseline model that should have been more specific.

\begin{figure}
\small
    \begin{tabular}{l}
    \toprule
    \vspace{.5\baselineskip}
    
    \begin{minipage}[h]{0.9\columnwidth}%
    \textbf{Context:} \ldots Rei et al. (2016) extended this model to include character embeddings in order to capture morphological similarities such as word endings. \textbf{[MASK]} \ldots \\ % At the same time, investigated the effectiveness of a number of auxiliary (morpho-syntactic) training objectives for the task of GED \ldots \\

	\textbf{Ground truth:} Rei (2017) \textit{subsequently} added a secondary LM objective to the neural sequence labeling architecture, operating on both word and character-level embeddings. This was found to be particularly useful for GED -introducing an LM objective allows the network to learn more generic features about language and composition.\\

	\textbf{Baseline:} Rei (2017) proposed a semi-supervised approach to the task of GED, using a language modeling objective to predict surrounding words for every word in the sentence. \\ 
	
	\textbf{Contexualized:} Rei (2017) \textit{further extended} this model with a secondary training objective, learning to predict the surrounding words in the context.\\
	\end{minipage}\tabularnewline
	\bottomrule
    \end{tabular}
    \caption{Example where the contextualized model correctly describes one reference paper as an extension of another (in \textit{italics}).}
    \label{tab:para_example_extend_one}
\end{figure}
    
\paragraph{The reference paper is an extension of another work.} When describing the historical development of a task or topic, it is common to present works in chronological order, where each work builds on top of the previous ones. A good citation generation model should recognize when the reference paper is an extension of another paper cited earlier in the context. Figure \ref{tab:para_example_extend_one} shows an example where the context discusses a paper, and the model needs to generate a citation for a follow-up work by the same first author. The contextualized model generates the phrase ``further extended" (referring to the earlier work, which already ``extended").

We observe that citations often contain an action verbs, such as \textit{extend}, and \textit{improve}, which are used when the reference paper is dependent on an earlier work. Compared with the baseline model, our contextualized model more frequently such verbs.

\section{Conclusion}

We present a simple reframing of the citation text generation task to make better use of citing context information. Our approach of generating the entire citing context window, with the target citation filled in, produces output that is more appropriate to its context than the existing method of generating the target citation alone. Our proposed approach changes only the generation target and is agnostic to any special features or input representations, so it is straightforward to apply to any existing citation generation models via retraining. Our human evaluation reveals that readers are able to judge when a citation better fits its context, and we present a qualitative analysis of some common shortcomings of citations generated by the baseline approach that are addressed by our contextualized model. 

\section*{Limitations}

The standard automatic evaluation metric for text generation, ROUGE, does not capture coherence well. This is a known problem: small differences in ROUGE do not correspond to noticeable differences in generation quality \cite{deutsch2022re}, and ROUGE does not capture other aspects of the summary, such as informativeness \cite{schluter2017limits} and factuality \cite{wallace2021generating}. We instead present a human evaluation and qualitative analysis to compare the coherence citations generated using our proposed approach with those of the baseline model. However, such evaluations are be expensive and time-consuming to conduct, requiring significant reading and cognitive effort from domain-expert judges; as a result, we are not able to evaluate a very large number of samples.

In addition, our work is focused on natural language processing papers published in the English language. Consequently, the scope of our model may not encompass the full range of diversity exhibited by papers from other fields (e.g., biology) or different sub-fields within computer science.

\section*{Ethics Statement}

Our work targets the task of citation generation, where plagiarism of the cited paper is a significant concern. We not attempt to control the level of extractiveness of our generated citations; as a result, they may copy extensively from the cited paper abstract, which given as input. 

Further, our approach generates citations that criticize the reference paper when it seems appropriate to do so based on the citation context. Errors or hallucinations by our model could result in false or unfair criticisms, which can mislead the reader and negatively impact their perception of the reference paper.

\section{Bibliographical References}\label{sec:reference}

\bibliographystyle{lrec-coling2024-natbib}
\bibliography{anthology,lrec-coling2024-example}

% \section{Language Resource References}
% \label{lr:ref}
% \bibliographystylelanguageresource{lrec-coling2024-natbib}
% \bibliographylanguageresource{languageresource}

\end{document}